# Learning to Distill:
# The Essence Vector Modeling Framework

**Kuan-Yu Chen**
Academia Sinica
Taipei, Taiwan
*kychen@iis.sinica.edu.tw*

**Shih-Hung Liu**
Academia Sinica
Taipei, Taiwan
*journey@iis.sinica.edu.tw*

**Berlin Chen**
National Taiwan Normal University
Taipei, Taiwan
*berlin@csie.ntnu.edu.tw*

**Hsin-Min Wang**
Academia Sinica
Taipei, Taiwan
*whm@iis.sinica.edu.tw*

## Abstract

In the context of natural language processing, representation learning has emerged as a newly active research subject because of its excellent performance in many applications. Learning representations of words is a pioneering study in this school of research. However, paragraph (or sentence and document) embedding learning is more suitable/reasonable for some tasks, such as sentiment classification and document summarization. Nevertheless, as far as we are aware, there is relatively less work focusing on the development of unsupervised paragraph embedding methods. Classic paragraph embedding methods infer the representation of a given paragraph by considering all of the words occurring in the paragraph. Consequently, those stop or function words that occur frequently may mislead the embedding learning process to produce a misty paragraph representation. Motivated by these observations, our major contributions in this paper are twofold. First, we propose a novel unsupervised paragraph embedding method, named the essence vector (EV) model, which aims at not only distilling the most representative information from a paragraph but also excluding the general background information to produce a more informative low-dimensional vector representation for the paragraph. We evaluate the proposed EV model on benchmark sentiment classification and multi-document summarization tasks. The experimental results demonstrate the effectiveness and applicability of the proposed embedding method. Second, in view of the increasing importance of spoken content processing, an extension of the EV model, named the denoising essence vector (D-EV) model, is proposed. The D-EV model not only inherits the advantages of the EV model but also can infer a more robust representation for a given spoken paragraph against imperfect speech recognition. The utility of the D-EV model is evaluated on a spoken document summarization task, confirming the practical merits of the proposed embedding method in relation to several well-practiced and state-of-the-art summarization methods.

## 1    Introduction

Representation learning has gained significant interest of research and experimentation in many machine learning applications because of its remarkable performance. When it comes to the field of natural language processing (NLP), word embedding methods can be viewed as pioneering studies (Bengio et al., 2003; Mikolov et al., 2013; Pennington et al., 2014). The central idea of these methods is to learn continuously distributed vector representations of words using neural networks, which seeks to probe latent semantic and/or syntactic cues that can in turn be used to induce similarity measures among words. A common thread of leveraging word embedding methods to NLP-related tasks is to represent a given paragraph (or sentence and document) by simply taking an average over the word embeddings corresponding to the words occurring in the paragraph. By doing so, this thread of methods has recently enjoyed substantial success in many NLP-related tasks (Collobert and Weston, 2008; Tang et al., 2014; Kageback et al., 2014).

Although the empirical effectiveness of word embedding methods has been proven recently, the composite representation for a paragraph (or sentence and document) is a bit queer. Theoretically, paragraph-based representation learning is expected to be more suitable for such tasks as information retrieval, sentiment analysis and document summarization (Huang et al., 2013; Le and Mikolov, 2014; Palangi et al., 2015), to name but a few. However, to the best of our knowledge, unsupervised paragraph embedding has been largely under-explored on these tasks. Classic paragraph embedding methods infer the representation of a given paragraph by considering all of the words occurring in the paragraph. Consequently, those stop or function words that occur frequently in the paragraph may mislead the embedding learning process to produce a misty paragraph representation. In other words, the frequent words or modifiers may overshadow the indicative words, thereby drifting the main theme of the semantic content in the paragraph. As a result, the learned representation for the paragraph might be undesired. In order to address this shortcoming, we propose a novel unsupervised paragraph embedding method, named the essence vector (EV) model, which aims at not only distilling the most representative information from a paragraph but also excluding the general background information to produce a more informative and discriminative low-dimensional vector representation for the paragraph.

On a separate front, with the popularity of the Internet and the increasing development of the digital storage capacity, unprecedented volumes of multimedia information, such as broadcast news, lecture recordings, voice mails and video streams, among others, have been quickly disseminated around the world and shared among people. Consequently, spoken content processing has become an important and urgent demand (Lee and Chen, 2005; Ostendorf, 2008; Liu and Hakkani-Tur, 2011). Obviously, speech is one of the most important sources of information about multimedia (Furui et al., 2012). A common school of processing multimedia is to transcribe the associated spoken content into text or lattice format by an automatic speech recognizer. After that, well-developed text processing frameworks can then be readily applied. However, such imperfect transcripts usually limit the associated efficacy. To bridge the performance gap between perfect and imperfect transcripts, we hence extend the proposed essence vector model to a denoising essence vector (D-EV) model, which not only inherits the advantages of the EV model but also can infer a more robust representation for a given spoken paragraph that is more resilient to imperfect speech recognition.

The remainder of this paper is organized as follows. We first briefly review two classic paragraph embedding methods in Section 2. Section 3 sheds light on our proposed essence vector model and its extension, the denoising essence vector model. Then, a series of experiments are presented in Section 4 to evaluate the proposed representation learning methods. Finally, Section 5 concludes the paper.

## 2  Literature Review

In contrast to the large body of work on developing various word embedding methods, there are relatively few studies concentrating on learning paragraph representations in an unsupervised manner (Huang et al., 2013; Le and Mikolov, 2014; Chen et al., 2014; Palangi et al., 2015). Representative methods include the distributed memory model (Le and Mikolov, 2014) and the distributed bag-of-words model (Le and Mikolov, 2014; Chen et al., 2014).

### 2.1  The Distributed Memory Model

The distributed memory (DM) model is inspired and hybridized from the traditional feed-forward neural network language model (NNLM) (Bengio et al., 2003) and the recently proposed word embedding methods (Mikolov et al., 2013). Formally, given a sequence of words, $\{w^1, w^2, \cdots, w^L\}$, the objective function of feed-forward NNLM is to maximize the total log-likelihood,

$$\sum_{l=1}^{L} \log P(w^l | w^{l-n+1}, \cdots, w^{l-1}). \quad (1)$$

Obviously, NNLM is designed to predict the probability of a future word, given its $n-1$ previous words. The input of NNLM is a high-dimensional vector, which is constructed by concatenating (or taking an average over) the word representations of all words within the context (i.e., $w^{l-n+1}, \cdots, w^{l-1}$), and the output can be viewed as that of a multi-class classifier. By doing so, the $n$-gram probability can be calculated through a softmax function at the output layer:

$$P(w^l | w^{l-n+1}, \cdots, w^{l-1}) = \frac{\exp(y_{w^l})}{\sum_{w_i \in V} \exp(y_{w_i})}, \quad (2)$$

where $y_{w_i}$ denotes the output value for word $w_i$, and $V$ is the vocabulary.

Based on the NNLM, the notion underlying the DM model is that a given paragraph also contributes to the prediction of the next word, given its previous words in the paragraph (Le and Mikolov, 2014). To make the idea work, the training objective function is defined by

$$\sum_{t=1}^{T} \sum_{l=1}^{L_t} \log P(w^l | w^{l-n+1}, \cdots, w^{l-1}, D_t), \quad (3)$$

where T denotes the number of paragraphs in the training corpus, $D_t$ denotes the $t$-th paragraph, and $L_t$ is the length of $D_t$. Since the model acts as a memory unit that remembers what is missing from the current context, it is named the distributed memory (DM) model.

## 2.2 The Distributed Bag-of-Words Model

Opposite to the DM model, a simplified version is to only rely on the paragraph representation to predict all of the words occurring in the paragraph (Le and Mikolov, 2014; Chen et al., 2014). The training objective function can then be defined by maximizing the predictive probabilities all over the words occurring in the paragraph:

$$\sum_{t=1}^{T} \sum_{l=1}^{L_t} \log P(w^l | D_t). \quad (4)$$

Since the simplified model ignores the contextual words at the input layer, the model is named the distributed bag-of-words (DBOW) model. In addition to being conceptually simple, the DBOW model only needs to store the softmax weights, whereas the DM model stores both softmax weights and word vectors (Le and Mikolov, 2014).

# 3 Learning to Distill

## 3.1 The Essence Vector Model

Classic paragraph embedding methods infer the representation of a paragraph by considering all of the words occurring in the paragraph. However, we all agree upon that the number of content words in a paragraph is usually less than that of stop or function words. Accordingly, those stop or function words may mislead the representation learning process to produce an ambiguous paragraph representation. In other words, the frequent words or modifiers may overshadow the indicative words, thereby making the learned representation deviate from the main theme of the semantic content expressed in the paragraph. Consequently, the associated capacity will be limited. In order to complement such deficiency, we hence strive to develop a novel unsupervised paragraph embedding method, which aims at not only distilling the most representative information from a paragraph but also diminishing the impact of the general background information (probably predominated by stop or function words), so as to deduce an informative and discriminative low-dimensional vector representation for the paragraph. We henceforth term this novel unsupervised paragraph embedding method the essence vector (EV) model.

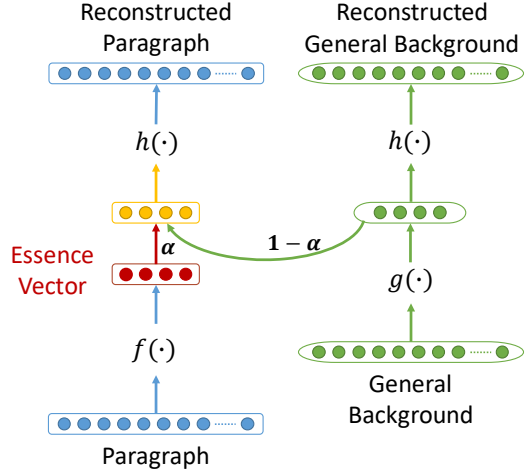

Figure 1: Illustrations of the essence vector model.

To turn the idea into a reality, we begin with an assumption that each paragraph (or sentence and document) can be assembled by two components: the paragraph specific information and the general background information. This assumption also holds in the low-dimensional representation space. Accordingly, the proposed method consists of three modules: a paragraph encoder $f(\cdot)$, which can automatically infer the desired low-dimensional vector representation by considering only the paragraph-specific information; a background encoder $g(\cdot)$, which is used to map the general background information into a low-dimensional representation; and a decoder $h(\cdot)$ that can reconstruct the original paragraph by combining the paragraph representation and the background representation.

More formally, given a set of training paragraphs $\{D_1, \cdots, D_t, \cdots, D_T\}$, in order to modulate the effect of different lengths of paragraphs, each paragraph is first represented by a bag-of-words high-dimensional vector $P_{D_t} \in \mathbb{R}^{|V|}$, where each element corresponds to the frequency count of a word/term in the vocabulary $V$, and the vector is normalized to unit-sum. Then, a paragraph encoder is applied to extract the most specific information from the paragraph and encapsulate it into a low-dimensional vector representation:

$$f(P_{D_t}) = v_{D_t}. \qquad (5)$$

At the same time, the general background is also represented by a high-dimensional vector with normalized word/term frequency counts, $P_{BG} \in \mathbb{R}^{|V|}$, and a background encoder is used to compress the general background information into a low-dimensional vector representation:

$$g(P_{BG}) = v_{BG}. \qquad (6)$$

Both $f(\cdot)$ and $g(\cdot)$ are fully connected deep networks with different model parameters $\theta_f$ and $\theta_g$, respectively. It is worthy to note that $f(\cdot)$ and $g(\cdot)$ can have same or different architectures. Since each learned paragraph representation $v_{D_t}$ only contains the most informative/discriminative part of $P_{D_t}$, we assume that the weighted combination of $v_{D_t}$ and $v_{BG}$ can be mapped back to $P_{D_t}$ by a decoder $h(\cdot)$:

$$h(\alpha_{D_t} \cdot v_{D_t} + (1 - \alpha_{D_t}) \cdot v_{BG}) = P'_{D_t}, \qquad (7)$$

where $h(\cdot)$ is also a fully connected multilayer neural network with parameter $\theta_h$, and the interpolation weight can be determined by an attention function $q(\cdot,\cdot)$:

$$\alpha_{D_t} = q(v_{D_t}, v_{BG}). \qquad (8)$$

The attention function can be realized by a trainable network or a simple linear/non-linear function. Further, to ensure the quality of the learned background representation $v_{BG}$, it should also be mapped back to $P_{BG}$ by $h(\cdot)$ appropriately:

$$h(v_{BG}) = P'_{BG}. \qquad (9)$$

In a nutshell, the training objective function of the proposed essence vector model is to minimize the total KL-divergence measure:

$$\min_{\theta_f, \theta_g, \theta_h} \sum_{t=1}^{T} \left( P_{D_t} \log \frac{P_{D_t}}{P'_{D_t}} + P_{BG} \log \frac{P_{BG}}{P'_{BG}} \right). \qquad (10)$$

The activation function used in the EV model is the hyperbolic tangent, except that the output layer in the decoder $h(\cdot)$ is the softmax (Goodfellow et al., 2016), the cosine distance is used to calculate the attention coefficients, and the Adam (Kingma and Ba, 2015) is employed to solve the optimization problem. At test time, a given paragraph can obtain its own representation by being passed through the paragraph encoder (i.e., $f(\cdot)$). Figure 1 illustrates the architecture of the EV model.

### 3.2 The Denoising Essence Vector Model

Next, we turn to focus on learning representations for spoken paragraphs. In addition to the stop/function words and modifiers, the additional challenge facing spoken paragraph learning is the imperfect transcripts generated by automatic speech recognition. Therefore, our goal is not only to inherit the advantages of the EV model, but also to infer a more robust representation for a given spoken paragraph that withstands the errors of imperfect transcripts. The core idea is that the learned representation of a spoken paragraph should be able to interpret its corresponding manual transcript paragraph as much as possible. With the intention of equipping the ability that can distill the *true* information from a given *spoken* paragraph, we further incorporate a multi-task learning strategy in the EV modeling framework. To put the idea into a reality, an additional module, a denoising decoder $s(\cdot)$, is introduced on top of the EV model. More formally, given a set of training spoken paragraphs $\{D_1, \cdots, D_t, \cdots, D_T\}$ and their manual transcripts $\{D_1^m, \cdots, D_t^m, \cdots, D_T^m\}$, the EV model can first be constructed by referring to each pair of $D_t$ and the general background information (cf. Section 3.1). Since we target at making the learned paragraph representation $v_{D_t}$ contain the true information of $D_t^m$, we assume that the weighted combination of $v_{D_t}$ and $v_{BG}$ can also be well mapped back to $P_{D_t^m}$ by the decoder $s(\cdot)$:

$$s\left(\alpha_{D_t} \cdot v_{D_t} + (1 - \alpha_{D_t}) \cdot v_{BG}\right) = P'_{D_t^m}, \qquad (11)$$

where $s(\cdot)$ is a fully connected neural network with parameter $\theta_s$. The activation function used in $s(\cdot)$ is the hyperbolic tangent, except that the last layer is the softmax. We will henceforth term this extended unsupervised paragraph embedding method the denoising essence vector (D-EV) model. The training objective of the D-EV model is to minimize the following total KL-divergence measure:

$$\min_{\theta_f, \theta_g, \theta_h, \theta_s} \sum_{t=1}^{T} \left( P_{D_t} \log \frac{P_{D_t}}{P'_{D_t}} + P_{D_t^m} \log \frac{P_{D_t^m}}{P'_{D_t^m}} + P_{BG} \log \frac{P_{BG}}{P'_{BG}} \right). \qquad (12)$$

## 4 Experimental Setup & Results

### 4.1 Experiments on the EV Model for Sentiment Analysis

At the outset, we evaluate the proposed EV model on the sentiment polarity classification task. Four widely-used benchmark multi-domain sentiment datasets are used in this study[1] (Blitzer et al., 2007). They are product reviews taken from Amazon.com in four different domains: Books, DVD, Electronics, and Kitchen. Each of the reviews, ranging from Star-1 to Star-5, were rated by a customer. The reviews with Star-1 and Star-2 were labelled as Negative, and those with Star-4 and Star-5 were labeled as Positive. Each of the four datasets contains 1,000 positive reviews, 1,000 negative reviews, and a number of unlabeled reviews. Labeled reviews in each domain are randomly split up into ten folds (with nine folds serving as the training set and the remaining one as the test set). All of the following results are reported in terms of an average accuracy of ten-fold cross validation. The linear kernel SVM (Chang and Lin, 2011) is used as our classifier and all of the parameters are set to the default values. All of the unlabeled reviews are used to obtain the general background information and train the EV model.

---
[1] https://www.cs.jhu.edu/~mdredze/datasets/sentiment/

|                        | Books | DVD   | Electronics | Kitchen | Average |
|------------------------|-------|-------|-------------|---------|---------|
| PCA                    | 0.762 | 0.769 | 0.807       | 0.824   | 0.790   |
| EV                     | **0.796** | **0.812** | **0.839** | **0.858** | **0.826** |
| Unigrams               | 0.797 | 0.805 | 0.837       | 0.860   | 0.824   |
| Bigrams                | 0.798 | 0.779 | 0.819       | 0.857   | 0.813   |
| Unigrams+Bigrams       | 0.810 | 0.821 | 0.852       | 0.884   | 0.842   |
| Unigrams+PCA           | 0.799 | 0.812 | 0.835       | 0.860   | 0.826   |
| Unigrams+EV            | 0.806 | 0.813 | 0.833       | 0.871   | 0.831   |
| Unigrams+Bigrams+PCA   | 0.810 | 0.821 | 0.852       | 0.884   | 0.842   |
| Unigrams+Bigrams+EV    | **0.838** | **0.824** | **0.862** | **0.890** | **0.853** |

Table 1: Experimental results on sentiment analysis achieved by the proposed EV model and other baseline features, including unigrams, bigrams, PCA, and the combinations.

In this set of experiments, we first compare the EV model with PCA (Bengio et al., 2013), which is a standard dimension reduction method. It is worthy to note that PCA is a variation of an auto-encoder (Bengio et al., 2013) method; thus it can be treated as our baseline system. All of the experimental results are listed in Table 1. As expected, the proposed EV model consistently outperforms PCA in every domain by a significant margin. The reason might be that PCA maps data to a low-dimensional space by maximizing the statistical variance of data, but the implicitly denoising strategy and the linear formulation limit its model capability. On the contrary, the proposed EV model is designed to distill the most useful information from a given paragraph and exclude the general background information explicitly; it thus can deduce a more informative and discriminative representation.

Next, we make a step forward to compare the EV model with other baseline systems based on literal bag-of-words features, including unigrams and bigrams. The results are also shown in Table 1. Several observations can be drawn from the results. First, although bigram features (denotes as Bigrams in Table 1) are believed to be more discriminative than unigram features (denotes as Unigrams in Table 1), the results indicate that Unigrams outperform Bigrams in most cases. The reason might be probably due to the curse of dimensionality problem. Second, as expected, the combination of unigram and bigram features (denotes as Unigrams+Bigrams) achieves better results than using Unigrams and Bigrams in isolation for all cases. Third, both the proposed EV model and PCA can make further performance gains when paired with Unigrams, Bigrams, and their combination. Fourth, the proposed EV model demonstrates its ability in the sentiment classification task since it consistently outperforms PCA for all cases in the experiments.

### 4.2 Experiments on the EV Model for Multi-Document Summarization

We further investigate the capability of the EV model on an extractive multi-document summarization task. In this study, we carry out the experiments with the DUC 2001, 2002, and 2004 datasets[2]. All the documents were compiled from newswires, and were grouped into various thematic clusters. The summary length was limited to 100 words for both DUC 2001 and DUC 2002, and 665 bytes for DUC 2004. The general background information was inferred from the LDC

---

[2] http://www-nlpir.nist.gov/projects/duc/

|  |  | ROUGE-1 | ROUGE-2 |
|---|---|---|---|
| 2001 | Peer T | 0.330 | 0.079 |
|  | VSM | 0.286 | 0.049 |
|  | LexRank | 0.334 | 0.061 |
|  | EV | 0.332 | 0.059 |
|  | CNN | 0.352 | 0.076 |
|  | PriorSum | 0.360 | 0.079 |
| 2002 | Peer 26 | 0.352 | 0.076 |
|  | VSM | 0.304 | 0.056 |
|  | LexRank | 0.353 | 0.075 |
|  | EV | 0.354 | 0.074 |
|  | CNN | 0.357 | 0.087 |
|  | PriorSum | 0.366 | 0.090 |
| 2004 | Peer 65 | 0.379 | 0.092 |
|  | VSM | 0.337 | 0.072 |
|  | LexRank | 0.379 | 0.089 |
|  | EV | 0.376 | 0.084 |
|  | CNN | 0.379 | 0.099 |
|  | PriorSum | 0.389 | 0.101 |

Table 2: Experimental results of multi-document summarization achieved by the proposed EV model and several state-of-the-art summarization methods.

Gigaword corpus[3] (including Associated Press Worldstream (AP), New York Times Newswire Service (NYT), and Xinhua News Agency (XIN)). The most common belief in the document summarization community is that relevance and redundancy are two key factors for generating a concise summary. In this paper, we leverage a density peaks clustering summarization method (Rodriguez and Laio, 2014; Zhang et al., 2015), which can take both relevance and redundancy information into account at the same time. That is, a concise summary for a given document set can be automatically generated through a one-pass process instead of an iterative process. Recently, the summarization method has proven its empirical effectiveness (Zhang et al., 2015). For evaluation, we adopt the widely-used automatic evaluation metric ROUGE (Lin, 2003), and take ROUGE-1 and ROUGE-2 (in F-scores) as the main measures following Cao et al., (2015).

We compare the proposed EV model with two baseline systems (the vector space model (VSM) (Gong and Liu, 2001) and the LexRank (Erkan and Radev, 2004) method), the best peer systems (including Peer T, Peer 26, and Peer 65) participating DUC evaluations, and the recently elaborated DNN-based systems (including CNN and PriorSum) (Cao et al., 2015). Owing to the space limitation, we omit the detailed introduction to these summarization methods; interested readers may refer to Penn and Zhu (2008), Liu and Hakkani-Tur (2011), Nenkova and McKeown (2011), and Cao et al., (2015) for more in-depth elaboration. It is worthy to note that the proposed EV model, the two baseline systems, and the best peer systems are unsupervised methods, while the DNN-based systems are supervised ones. The experimental results are listed in Table 2. Several interesting observations can be concluded from the results. First, the proposed EV model outperforms VSM by a large margin in all cases, and performs comparably to other well-designed unsupervised summarization methods. Second, both LexRank and EV (with the density peaks clustering method) take pairwise information into account globally, so their results are almost the same. Third, although the proposed EV model is an unsupervised method and is not specifically designed toward summarization, it almost achieves the same performance level as the complicated DNN-based supervised methods (i.e., CNN and PriorSum), which confirms the power of the EV model again.

---

[3] https://catalog.ldc.upenn.edu/LDC2011T07

|  | ROUGE-1 | ROUGE-2 | ROUGE-L |
|---|---|---|---|
| DM | 0.387 | 0.242 | 0.337 |
| DBOW | 0.396 | 0.250 | 0.344 |
| EV | **0.414** | 0.264 | 0.361 |
| D-EV | **0.414** | **0.278** | **0.374** |
| MRW | 0.332 | 0.191 | 0.291 |
| LexRank | 0.305 | 0.146 | 0.254 |
| SM | 0.332 | 0.204 | 0.303 |
| ILP | 0.348 | 0.209 | 0.306 |

Table 3: Experimental results of spoken document summarization achieved by the proposed EV and D-EV models and several state-of-the-art summarization methods.

### 4.3 Experiments on the D-EV Model for Spoken Document Summarization

In order to assess the utility of the proposed D-EV model, we perform a series of experiments on the extractive spoken document summarization task. All of experiments are conducted on a Mandarin benchmark broadcast new corpus[4] (Wang et al., 2005). The MATBN dataset is publicly available and has been widely used to evaluate several NLP-related tasks, including speech recognition (Chien, 2015), information retrieval (Huang and Wu, 2007) and summarization (Liu et al., 2015). As such, we follow the experimental setting used in previous studies for speech summarization in the literature. The vocabulary size is about 72 thousand words. The average word error rate of the automatic transcripts of these broadcast news documents is about 38%. The reference summaries were generated by ranking the sentences in the manual transcript of a broadcast news document by importance without assigning a score to each sentence. Each document has three reference summaries annotated by three subjects. For the assessment of summarization performance, we adopt the commonly-used ROUGE metric (Lin, 2003), and take ROUGE-1, ROUGE-2 and ROUGE-L (in F-scores) as the main measures. The summarization ratio is set to 10%. An external set of about 100,000 text news documents, which was assembled by the Central News Agency (CNA) during the same period as the broadcast news documents to be summarized (extracted from the Chinese Gigaword Corpus[5] released by LDC), is used to obtain the background representation.

To begin with, we compare the performance levels of the proposed EV and D-EV models and two classic paragraph embedding methods (i.e., DM and DBOW) for spoken document summarization. All the models are paired with the density peaks clustering summarization method. The results are shown in Table 3, from which several observations can be drawn. First, DBOW outperforms DM in our experiments, though DBOW is a simplified version of DM. Second, the proposed EV model outperforms DM and DBOW by a large margin, as expected. The results confirm that EV can modulate the impact of those stop or function words when inferring representations for paragraphs. That is to say, the proposed paragraph embedding method EV can indeed distill the most important aspects of a given paragraph and meanwhile suppress the impact of the general background information for producing a more discriminative paragraph representation. Thus, the relevance degree between any pair of sentence and document representations can be estimated more accurately. Third, the D-EV model consistently outperforms other paragraph embedding methods, including our own EV model. The outcome reveals that, although EV can achieve better performance than other classic paragraph embedding methods, the recognition errors inevitably make the inferred representations deviate from the original semantic content of spoken paragraphs. Accordingly, the results signal that the D-EV model can complement the deficiency of the EV model in spoken document summarization; we thus believe that it is more suitable for use in spoken content processing.

---

[4] http://slam.iis.sinica.edu.tw/corpus/MATBN-corpus.htm
[5] https://catalog.ldc.upenn.edu/LDC2011T13

In the last set of experiments, we compare the results mentioned above with that of several well-practiced, state-of-the-art unsupervised summarization methods, including the graph-based methods (i.e., the Markov random walk (MRW) method (Wan and Yang, 2008) and the LexRank method (Erkan and Radev, 2004)) and the combinatorial optimization methods (i.e., the submodularity-based (SM) method (Lin and Bilmes, 2010) and the integer linear programming (ILP) method (Riedhammer et al., 2010)). Among them, the ability of reducing redundant information has been aptly incorporated into the submodular-based method and the ILP method. Interested readers may refer to Penn and Zhu (2008), Liu and Hakkani-Tur (2011), and Nenkova and McKeown (2011) for comprehensive reviews and new insights into the major methods that have been developed and applied with good success to a wide range of spoken document summarization tasks. The results are also listed in Table 3. Several noteworthy observations can be drawn from the results of these methods. First, although the two graph-based methods (i.e., MRW and LexRank) have similar motivations, MRW outperforms LexRank by a large margin. Second, although both SM and ILP have the ability to reduce redundant information when selecting indicative sentences to form a summary for a given document, ILP consistently outperforms SM. The reason might be that ILP performs a global optimization process to select representative sentences, whereas SM chooses sentences with a recursive strategy. Comparing the results of these strong baseline systems to that of the paragraph embedding methods (including DM, DBOW, EV, and D-EV) paired with the density peaks clustering summarization method, it is clear that all the paragraph embedding methods are better than the baseline methods. The results corroborate that, instead of only considering literal term matching for determining the similarity degree between a pair of sentence and document, incorporating concept (semantic) matching into the similarity measure leads to better performance. In particular, the proposed D-EV model is the most robust among all the methods compared in the paper, which supports the important notion of the proposed "*learning to distilling*" framework. We also want to note that the proposed methods (i.e., EV and D-EV) can also be incorporated with the graph-based methods and the combinatorial optimization methods. We leave this exploration for future work.

## 5    Conclusions

In this paper, we have proposed a novel paragraph embedding framework, which is embodied with the essence vector (EV) model and the denoising essence vector (D-EV) model, and made a step forward to evaluate the proposed methods on benchmark sentiment classification and document summarization tasks. Experimental results demonstrate that the proposed framework is the most robust among all the methods (including several well-practiced or/and state-of-the-art methods) compared in the paper, thereby indicating the potential of the new paragraph embedding framework. For future work, we will first focus on pairing the (denoising) essence vector model with other summarization methods. Moreover, we will explore other effective ways to integrate extra cues, such as speaker identities and relevance information, into the proposed framework. Furthermore, we also plan to extend the applications of the proposed framework to information retrieval and language modeling, among others.